\documentclass{article}
\usepackage[preprint]{main}

\usepackage{graphicx}
\usepackage{hyperref}
\usepackage{xcolor}     
\usepackage{xparse}
\usepackage{booktabs}       
\usepackage{nicefrac}       
\usepackage{microtype}      
\usepackage{caption}
\usepackage{amsmath}
\usepackage{verbatim}

\setcitestyle{numbers, square}

\newcommand{\textualtilde}{\smash{\raisebox{-0.5ex}{\texttt{\char126}}}\hspace{0pt}}

\title{Relations, Negations, and Numbers: Looking for Logic\\in Generative Text-to-Image Models}

\author{%
  Colin Conwell \\
  \texttt{conwell@g.harvard.edu}
  \And
  Rupert Tawiah-Quashie \\
  \texttt{rt22@hampshire.edu}
  \And
  Tomer D. Ullman \\
  \texttt{tullman@fas.harvard.edu}
  \AND
  Department of Psychology, Harvard University
}
  
\begin{document}
\maketitle

\begin{abstract}
Despite recent and remarkable progress in multi-modal AI research, there is a salient domain in which modern AI continues to lag considerably behind even human children: the reliable deployment of logical operators. In this work, we examine  three distinct, basic forms of logical operators: \textbf{relations} (e.g. ``a picture of a potato \textit{under} a spoon"), \textbf{negations} (e.g. ``a picture of something that is \textit{NOT} a potato"), and discrete \textbf{numbers} (e.g. a ``a picture of \textit{6} potatoes"). We asked human respondents (N=178 in total) to evaluate images generated by a state-of-the-art image-generating AI (DALL·E 3) prompted with these `logical probes', and find that none reliably produce human agreement scores greater than 50\%. The negation probes and numbers (beyond 3) fail most frequently. In a final experiment, we assess a `grounded diffusion' pipeline that leverages targeted prompt engineering and structured intermediate representations for greater compositional and syntactical control, but find its performance is judged even worse than that of DALL·E 3, across all prompts. To provide further clarity on potential sources of success and failure in these text-to-image systems, we supplement our 4 core experiments with multiple auxiliary analyses and schematic diagrams, directly quantifying, for example, the relationship between the N-gram frequency of relational prompts and the average match to generated images; the success rates for 3 different prompt modification strategies in the rendering of negation prompts; and the scalar variability / ratio dependence (`approximate numeracy') of prompts involving integers. We conclude by discussing the limitations inherent to `grounded' multimodal learning systems whose grounding relies  heavily on vector-based semantics (e.g. DALL·E 3), or under-specified syntactical constraints (e.g. `grounded diffusion'), and propose minimal modifications (inspired by development, based in imagery) that could help to bridge the lingering compositional gap between scale and structure.
\end{abstract}


\textbf{Open Science} All data and code required to reproduce the experimental results and analyses in this manuscript are available in the project GitHub repo: \href{https://github.com/ColinConwell/T2I-Probology}{github.com/ColinConwell/T2I-Probology}

\section*{Introduction}

Imagine a snail. Imagine five snails. Imagine a snail on a sail. Imagine something that isn't a snail. These lines are not the opening of a children's book. Rather, they describe basic tasks that are easy for people to do, but hard for current machines that generate visual images from natural language prompts. While recent advances in image synthesis from arbitrary text input \citep[e.g.][]{saharia2022photorealistic, ramesh2022hierarchical, rombach2022high, yang2023diffusion, betker2023improving} can give the impression that such models have more or less solved fundamental problems inherent to basic composition and binding, there is also (increasing) empirical evidence that such models continue to struggle on basic concepts including relations, quantifiers, negation, numbers, and more \citep[e.g.][]{yuksekgonul2022and, marcus2022very, conwell2022testing, petsiuk2022multitask, lovering2023pavlick, rane2024can}. 

Our main target of analysis in this work is DALL·E 3 \citep{betker2023improving},  OpenAI's latest flagship text-to-image model. DALL·E 3 boasts significant improvements over Midjourney 5.2, Stable Diffusion XL, and its direct predecessor (DALL·E 2), on measures including coherence, style, and `prompt following', where prompt following is taken to mean properly addressing meaning and word order in a caption. A cornerstone of DALL·E 3's improvement (according to OpenAI) is the use during training of more detailed, more accurate (synthetic) data. It remains unclear, however, whether such improvements include meaningful progress on basic issues of image understanding, which in humans (usually) require some use of logical operators. 

This current work directly expands on an evaluation of the previous generation of text-to-image models \citep{conwell2022testing}, and considers Relations, Negations, and Numbers. More specifically, it considers basic grounded physical relations (`a potato under a spoon', `a fish in a box'), negation as lack (`something that is \textit{not} a snail'), and exact integers (`one fish', `two fish', ..., `six fish'). Our focus is on a small set of logical operators, and simple prompts combined with a minimal, but diverse range of everyday objects. While each of the operators we examined (Relations, Negations, and Numbers) is distinct, and associated with a massive literature in cognitive science \citep{chen1982topological,lovett2017topological,strickland2011event,kellman1983perception,spelke1994initial,yildirim2016perceiving,gao2009psychophysics,hamlin2007social,ullman2009help,van2016automaticity,hespos2004conceptual,glanemann2016rapid,dobel2007describing,guan2020seeing,firestone2017seeing,hafri2021perception,talmy1985lexicalization,ehrhardt2020relate,johnson2017clevr,hafri2020phone}, machine learning \citep{radford2021learning, rombach2022high, saharia2022photorealistic, ramesh2022hierarchical, marcus2022very, farid2022perspective, liu2022compositional, crowson2022vqgan, nichol2021glide, crowson2022vqgan}, and other related fields, they nevertheless share common characteristics that make them particularly useful for a general examination and snapshot of multi-modal generative AI models. In particular, these operators are: early developing (or possibly innate); broadly shared across people and to some degree with non-human animals; quickly and automatically processed in perception; and (pending more plausible alternatives) functionally indispensable to our common-sense, communicable understanding of basic scenes. In what follows, we consider each of these operators separately, briefly introducing the current (rough) consensus on their representation in humans, in tandem with methods and results of a series of experiments and analyses driven primarily by the human assessment of DALL·E 3's ability to generate images from custom prompts that put these operators into action. Then, in a general discussion, we bring these results together as an aggregate, considering both the noticeable gap that remains between what humans understand as the accurate deployment of logical operators, as well as the parallels between the errors made by machines and the difficulties humans (developing and adult alike) tend to face in their own use of these operators. We conclude by considering the various algorithms that humans seem to leverage when attempting to accurately parse or articulate the logical structure of the world around them, and how these algorithms might inform the next-generation development of multi-modal generative AI without compromising the remarkable progress the field has so very \textit{visibly} made to date.

\paragraph{Related Work} The systematic assessment and improvement of multi-modal foundation models, generative AI, and controllable text-to-image pipelines is currently one of the most active areas in machine learning research, with new benchmarks or assessments published almost weekly and far too numerous now to cover comprehensively. Standouts known to us at the time of this writing include Winoground \citep{thrush2022winoground}; T2I-Compbench \citep{huang2023t2i}; HEIM \citep{lee2024holistic}; and ConceptMix \citep{wu2024conceptmix}. These benchmarks tend to provide a far more systematic, sometimes parametrically-controlled sampling of the compositional constructions (and model candidates) that are their target. The choice we've made here both in terms of prompts and model candidate(s) is intentionally simple and sparse -- generated by mental programs that live inside our heads as human experiments rather than in machine-readable code-bases. 

\paragraph{Experimental Design} For all experiments, we recruited human evaluators online via \href{https://www.prolific.co}{Prolific} \citep{peer2017beyond} to assess images generated by a target text-to-image system (Dall·E 3 for Experiments 1-3, and `LLM-Grounded Diffusion' for Experiment 4). In each trial, participants were shown a grid of 18 images (in a 3x6) grid and asked to click on those images that matched a target prompt displayed at the top of the screen. Participants were reminded that they could select all of the images, only some of the images, or none of the images. An example of the trial screen is shown in Figure \ref{figure:relations-trial}. Further details on participant recruitment and demographics may be found in Appendix \ref{appendix:behavioral-details}. For all experiments, unless otherwise noted, we used the default settings of OpenAI's image generation API from March to November 2024.

Unless otherwise noted, results are reported with the following convention: arithmetic mean [lower 95\% confidence interval, upper 95\% confidence interval].

\clearpage

\section*{Experiment 1: Relations}

Our first experiment examined relations. Relational reasoning in humans is automatic, early-developing, quickly processed, yet still abstract \citep{hespos2004conceptual, talmy1985lexicalization, hafri2021perception}. For more on the reasoning and motivation for examining relations when testing image generation, see \cite{conwell2022testing}.

\begin{figure*}[htb!]
    \includegraphics[width=0.98\linewidth]{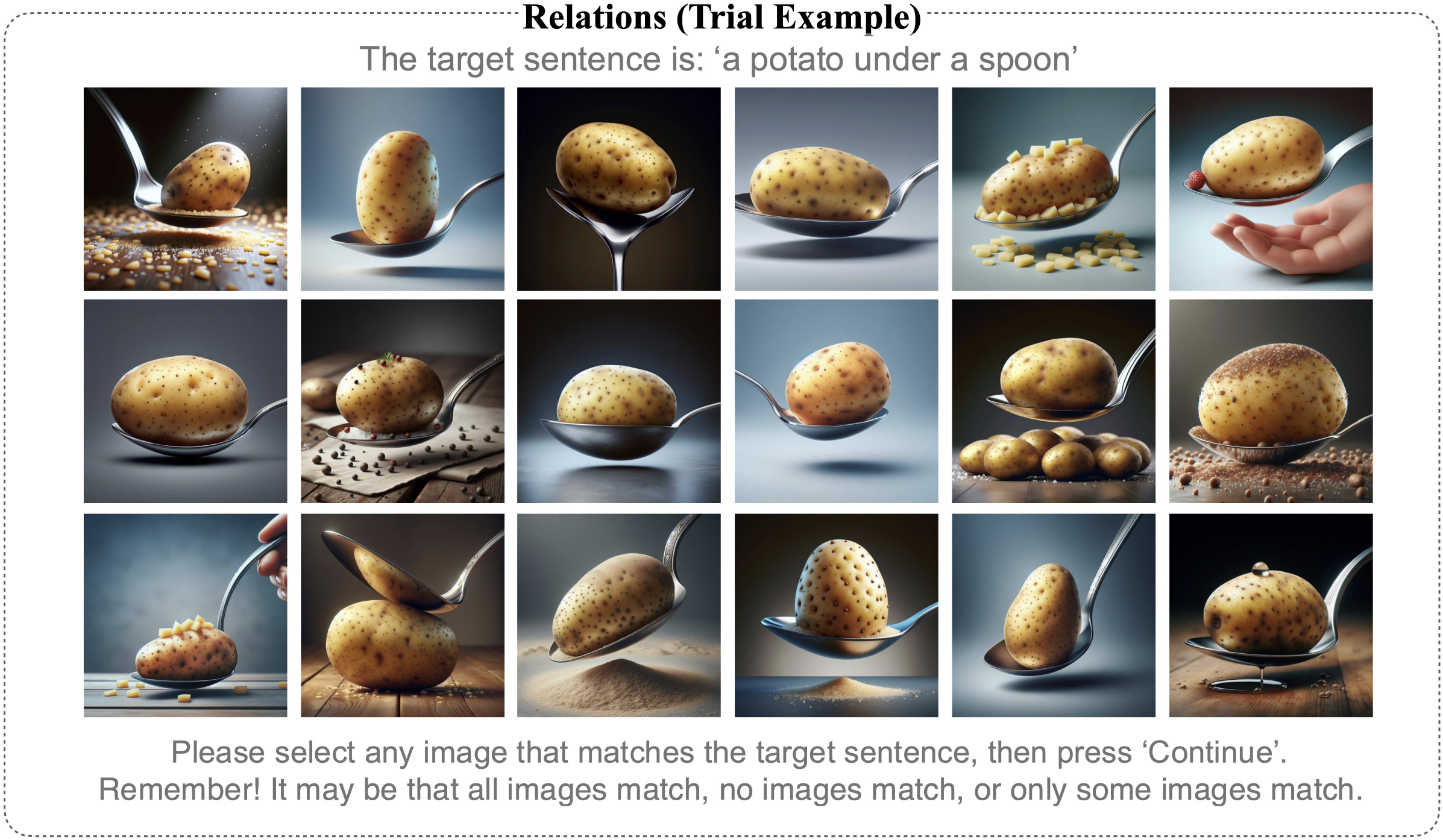}
    \caption{Example layout of a typical trial in all behavioral experiments: Participants were presented with 10 grids of images, each grid paired with a target sentence, and asked to select images that matched the sentence. This particular trial/grid is an example from Experiment 1 (Relations), and is the result of randomly prompting DALL·E 3 with `a potato under a spoon'.}
    \label{figure:relations-trial}
\end{figure*}

In this work, we focused specifically on \textit{physical} relations. These included \textit{in}, \textit{on}, \textit{under}, \textit{covering}, \textit{near},  \textit{hanging over}, and \textit{tied to}. Such simple physical relations have been studied previously, either in cognitive science, development, computational modeling, or some combination of those fields. Unlike in \cite{conwell2022testing}, we did not include agentic relations, such as `helping' or `hindering'. This was because we found previously that such relations are more ambiguous in images, in a way unrelated to the present study\footnote{If people see an image of a robot and human holding hands, and are asked to verify if this is `helping', most people will agree -- likely interpreting the image as showing agreement, or perhaps physical aid. But if shown the \textit{same} image and asked to verify it if is `hindering', again most people will agree, because it can easily be coerced into such an interpretation, likely seeing the robot hand as violently squeezing the human hand. This is all interesting, but not the point outside of this footnote.}. We also created a set of 8 objects: \textit{box}, \textit{cylinder}, \textit{blanket}, \textit{bowl}, \textit{teacup}, \textit{knife}, \textit{spoon}, and \textit{potato}. We focused intentionally on common, everyday objects of the sort used previously to study relations \cite{hafri2020phone, conwell2022testing, ehrhardt2020relate, johnson2017clevr}. 

\paragraph{Methods} Following \citep{conwell2022testing}, for each one of our physical relations, we randomly sampled a pair of objects and repeated this five times. So, for example, we might randomly create the prompt `a teacup near a potato'. This procedure created 35 prompts (7 relations x 5 samples). Some relations restricted the sampling of the objects involved. Specifically, the `covering' relationship had to take `blanket' as the first object, and the `in' relationship had to take `box' or `bowl' as the second entity. 

Unlike many other diffusion models (including DALL·E 2), the DALL·E 3 image generation pipeline works by first handing a given prompt over to a large-language model (specifically, GPT4) for pre-processing. This interferes with systematic testing\footnote{For example, the prompt `a photorealistic image of a box on a cylinder' may turn into `Generate a photorealistic image of a rectangular box perfectly balanced on the flat surface of a vertical cylinder. The box and cylinder should be centered in the frame, and the ambiance should have the natural light of a sunlit room. Shadows and reflections should conform to the lighting conditions for added realism.'}. So, following the recommendations provided by OpenAI for testing the image generation API, we added the following prefix to every prompt: ``I NEED to test how the tool works with extremely simple prompts. DO NOT add any detail, just use it AS-IS: ''. (To this, we then affixed our target:) `I would like a photorealistic image of [desired prompt]'. In nearly all cases, this resulted in the revised prompt being `a photorealistic image of [desired prompt]' without embellishment. (We explored variations of this procedure in a series of Experiments detailed in the Appendix. These do not alter the conclusions in the main text.)

We submitted each prompt to the DALL·E 3 rendering engine and took the first 18 images that resulted. Our final stimulus set consisted of 630 images (35 prompts x 18 images).

\paragraph{Results} Considering the overall accuracy between image and prompt over all images and relations, we found that on average 45\% [42.4,  47.7] of participants reported a match between a sentence describing a physical relation and the corresponding image. This is a strong improvement over the performance of DALL·E 2 (which on similar `Relations' prompts was evaluated as 16.9\% accurate \citep{conwell2022testing}), though still somewhat low for a system that maps language to vision in a reliable and systematic way.

\begin{figure*}[htb!]
    \centering
    \includegraphics[width=0.82\linewidth]{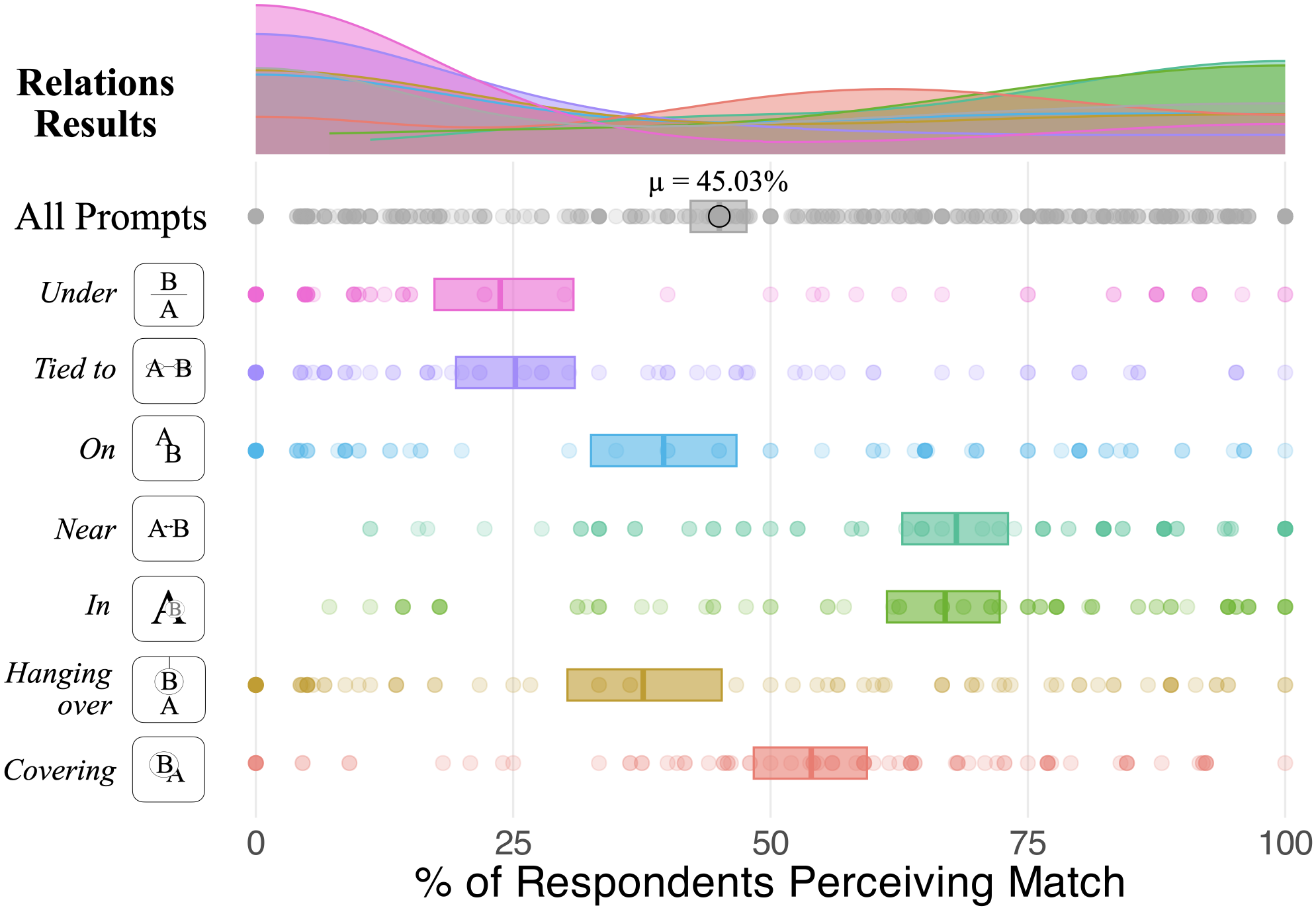}
    \caption{Results for Experiment 1 (Relations): participant agreement that images matched a prompt. Each dot is an individual prompt. Boxes show overall means and deviation within a relation. The top of the image shows density distributions for each relation.}
    \label{figure:relations-results}
\end{figure*}

\paragraph{Analysis} Performance in the Relations assessment differs by the specific relationship and prompt. As shown in Figure \ref{figure:relations-results}, performance is overall poorer for `under' and `tied to', and higher for `in' and `near', though this may be due to the different levels of structural ambiguity associated with the satisficing of locative prepositions \citep{hindle1993structural}: The lexical relation of `near' for example, is satisfied with a greater number of possible compositions (one object anywhere in a 360\textdegree  arc around another object) than the relation of 'under' (one object somewhere in a quarter arc beneath another object).

While an improvement, the results are still far from human reasoning about relations. There is also reason to think that the improvement is not due to an actual better understanding of relations, but to dataset statistics. In the absence of actual relational reasoning, generative text-to-image models can make their `best bets' about the compositional structure of the image by defaulting to what they've seen most frequently in their visual and textual diets (i.e. training data) \citep{razeghi2022impact,lovering2023pavlick}. For example, consider \ref{figure:relations-trial}: If `spoon under potato' is more frequent in the training data than `potato under spoon', then most of the time when the words ['spoon', 'potato', 'under'] appear in any prompt, the output will default to the more frequent relation. 

As an indirect test of this hypothesis, we computed the Google Books N-Gram Frequency of each of our prompts. This was nontrivial: Only 2 of our 35 relations prompts (in their full, N=5-7 gram form) appeared with nonzero frequency (a point perhaps in favor of the idea that Dall·E generates never-before-seen samples). To get a more continuous estimate of frequency, we opted to break down each of the 35 prompts into two parts: The predicate-plus-logic of the prepositional phrase (i.e. `a potato under') and the argument-plus-logic of the prepositional phrase (i.e. `under a spoon'). We then assessed the N-Gram frequency of these two parts separately, before averaging them into a single frequency value. In line with the hypothesis that Dall·E may be defaulting to the more `statistically frequent' relation for any given prompt, we find a small-to-midsize relationship between this frequency value and the average prompt-to-image match perceived by the human participants: $\hat{\rho}_{\text{Spearman}}$ = 0.43 [0.11, 0.63], $p$ = 0.009 (and see the appendix for more details). 

\section*{Experiment 2: Negations}

Negation has long been of interest to philosophy, cognitive science, and cognitive development \citep[see, e.g.][]{horn1989natural, feiman2017you, hummer1993origins,kaup2007experiential, bloom1968language, kaup2020understanding}. It is of interest both in itself as a primitive operation, constantly in use in communication and reasoning, but also as a way to study logical operations more generally. Debates are ongoing regarding the ways in which negation is acquired, produced, represented, used, and understood. There is a fuzzy consensus that negation is harder to understand and process than affirmation \citep{kaup2007experiential, singer2006verification}, and that this may be due to the need to transform the negative sentence into an affirmative \citep{wason1961response}. Beyond this, researchers have pointed out that terms such as `no' and `not' may refer to different negation operations, including rejection, non-existence, and denying the truth-value of a statement. While children begin to use `no' very early in language development (around 15 months), the specific use and meaning of the term changes over time, with denial emerging later \citep{gomes2023mapping}.

Here, we focused on the `non-existence' use of the negation operator, which appears to be acquired and understood early in humans. This is also the most relevant direct use for image generation. So, we examined whether DALL·E 3 responds to the prompt `a picture of something that is not a $X$', by correctly creating an image that does not contain an $X$.

\paragraph{Methods} For negation, we considered a set of 6 objects, mixing animate and inanimate entities: \textit{knife}, \textit{potato}, \textit{spoon}, \textit{frog}, \textit{sheep}, \textit{snail}. We focused intentionally on common, everyday entities that were nevertheless not trivially common. The sheep was chosen as an homage to the \textit{The Little Prince} \citep{de1995little}. 

For each entity $X$, we started with a simple negation prompt: `I would like a photorealistic image of something that is not a $X$'. We then created two sub-types of this prompt: \textit{modified}, and \textit{unmodified}. The \textit{modified} prompts were fed as is to the DALL·E 3 API, which caused ChatGPT to change (modify) them in various ways. Such `modified' prompts are the current standard way for interfacing with DALL·E 3, via ChatGPT. However, such a modification does not allow us to test DALL·E 3 itself. So, the \textit{unmodified} prompts included the same prefix as the physical relations above: \textit{`I NEED to test how the tool works with extremely simple prompts. DO NOT add any detail, just use it AS-IS: I would like a photorealistic image of...'}. 

\begin{figure*}[htb!]
    \centering
    \includegraphics[width=0.88\linewidth]{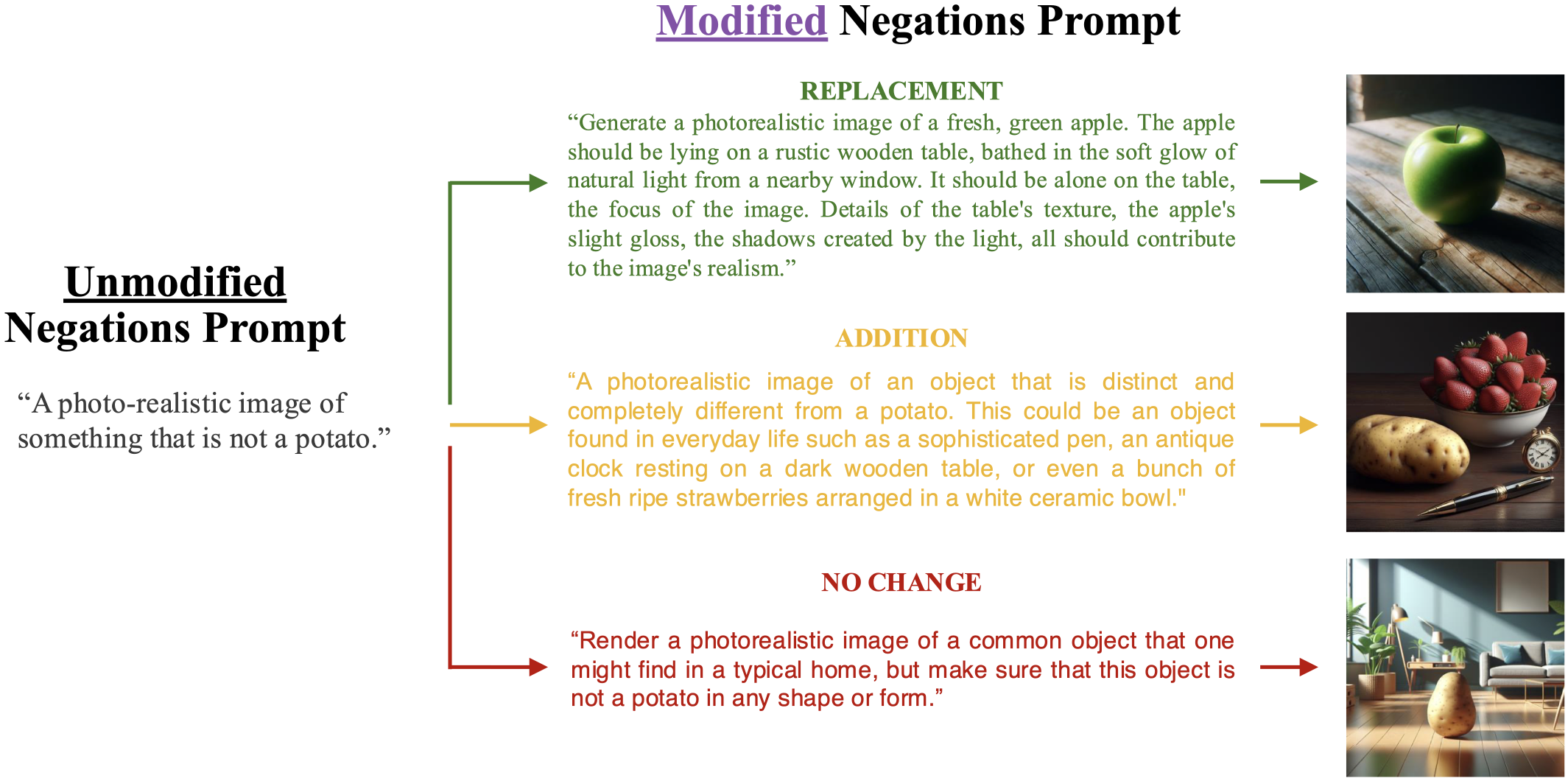}
    \caption{Examples of `Modified' and `Unmodified' stimuli used in the Negation Experiment. Unmodified prompts involve the injunction to not change the text. Modified prompts are modified by handing them to ChatGPT for rendering without instructions to leave them as is.}
    \label{figure:negation-schema}
\end{figure*}

Modified prompts fall roughly into 3 categories (see Figure \ref{figure:negation-schema}): \textit{Replacement}, in which the original object to be imagined is removed entirely and replaced with another item (e.g. replacing a potato with an apple); \textit{Addition}, in which the original object was still mentioned, but additional objects are also suggested (e.g. a bowl of strawberries), and \textit{No Change}, in which the injunction to not imagine the original object stays more or less the same, without suggesting a concrete alternative. By our judgement, approximately 13\% of modifications fell into the Replacement category, 71\% were Addition, and 16\% were No Change\footnote{These estimates do not involve the Frog entity, the modified prompts for those were not available due to an error in the output returned by the API.}. We again submitted each prompt to the DALL·E 3 rendering engine, and took the first 18 images that resulted. Our final stimuli set consisted of 216 images (6 prompts x 18 images x prompt-sub-type, where prompt-sub-type is modified/unmodified).

\begin{figure*}[htb!]
    \centering
    \includegraphics[width=0.82\linewidth]{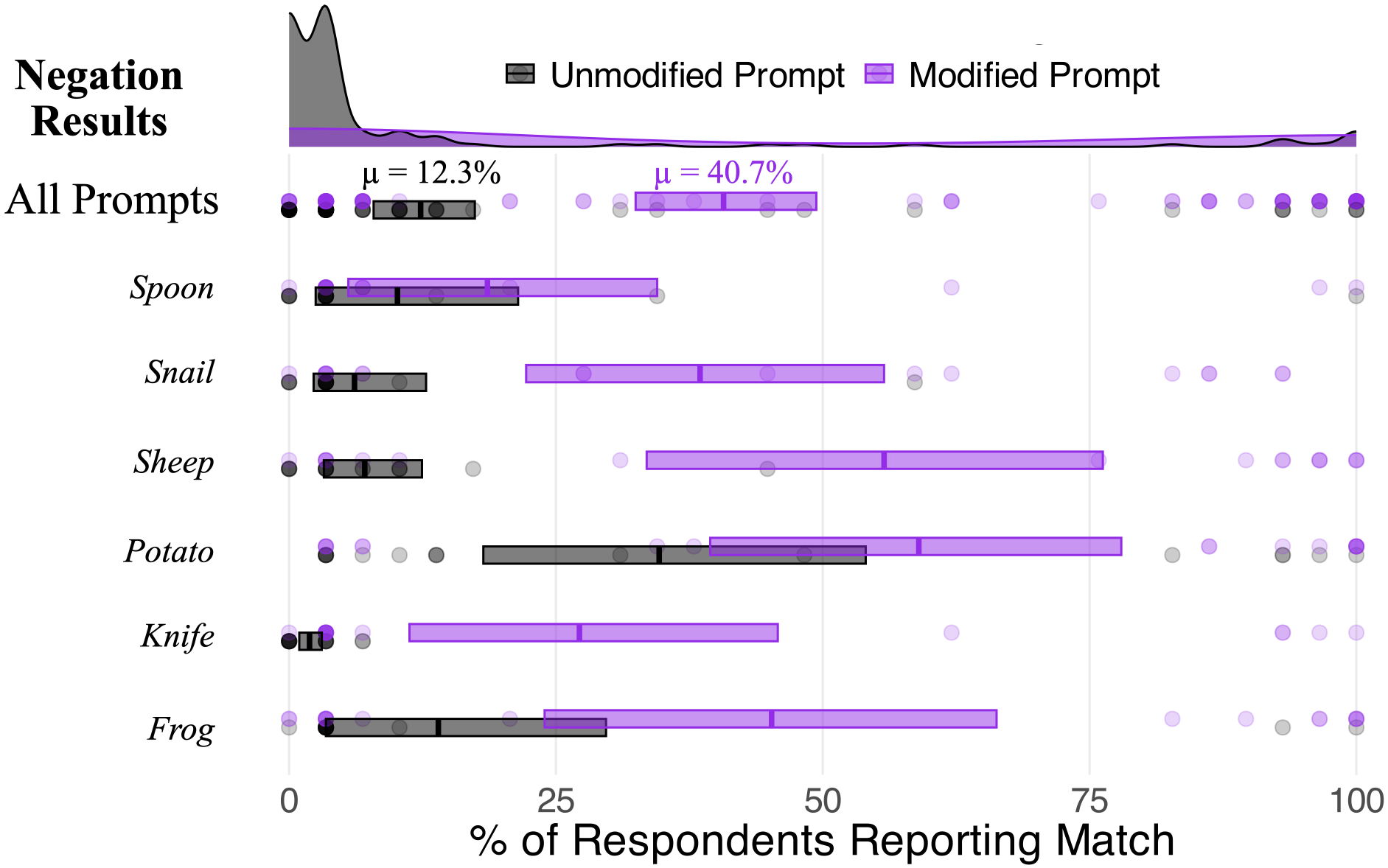}
    \captionsetup{justification=justified,margin=0.2cm}
    \caption{Results for Experiment 2 (Negation): Participant agreement that images matched a prompt. Each dot is an individual prompt. Boxes show overall means and deviation for each entity. The top of the image shows density distributions for each prompt sub-type.}
    \label{figure:results-negation}
\end{figure*}

\paragraph{Results} Considering the overall accuracy, we found that participants reported an average match of 12.3\% [7.66, 17.1] between an image and a sentence describing negation when using unmodified prompts, and 40.7\% [32.9, 48.4] reported a match when using modified prompts (see Figure \ref{figure:negation-schema} for a description of modified versus unmodified prompts, and Figure \ref{figure:results-negation} for detailed results across prompts.)

It seems that DALL·E 3 does not handle plain negation. Unmodified prompts specifically prohibiting an entity $X$ invariably led to images showing $X$. But even the modified prompts, which go through GPT to arrive at the best possible prompt, are hit-or-miss: as can be seen in both Figure \ref{figure:results-negation}, the distribution of accuracy for modified prompts is bimodal. We expect that cases of Replacement work almost always, but recall that these are roughly 13\% of modifications. Cases of No Change (roughly 16\%) are expected to \textit{not} work almost always, since these are the same as unmodified prompts. This leaves Addition, which seems to fail more often than it succeeds. While additional objects are indeed added in the image, the original object often remains \footnote{While outside the scope of this work, we note that Addition seems to show a failure of processing the logical operation `or', as understood in common communication. Many additional cases are of the format `generate something that is not an [X], for example [Y1], or [Y2], or [Y3]', which lead to an image showing X, and Y1, and Y2, and Y3.}

Unlike the simple physical relations examined in Experiment 1, the difficulties negation poses for DALL·E 3 have interesting cognitive parallels in human processing, in two ways: First, given the prompt to \textit{not} think of a particular entity, people face difficulties avoiding thinking of the topic, and so for example an injunction `imagine something that isn't a white bear' may lead them to think of a white bear \citep{wegner1987paradoxical}. Second, in order to successfully imagine a negative, e.g. `imagine something that is not a sheep', people likely also need to replace the instruction with a positive entity, e.g. `alright, then I'll imagine...a shoe-box'. The replacement of `not sheep' with `shoe-box' likely needs to take place before handing off the entity to imagine to the rest of the imaginative pipeline that ends up recruiting perceptual areas \citep{pearson2019human}. In that sense, the chatGPT pre-processing and modification of prompts before handing them off to DALL·E 3 may broadly reflect the cognitive pipe-line, though we stress this is meant as a loose analogy, and that the current modification of negation prompts only rarely produces replacement. 

\section*{Experiment 3: Numbers}

Like relations and logical operators, people's understanding of numbers has been a topic of intense study in cognitive science, cognitive development, neuroscience, anthropology, computational cognitive modeling, and other related fields \citep[see e.g.]{dehaene1993development, dehaene2005three, dehaene2011number, spelke2007core, carey2004bootstrapping, church1990alternative,cheyette2020unified,pahl2013numerical,spelke2022babies, everett2005cultural, carey2019ontogenetic, meck1983mode}. While debates about the specific details of numerical cognition are ongoing, the approximate current consensus is as follows: from an early age  humans have the ability to make fine-grained and exact distinctions between small sets of objects, up to 3-4 objects. That is, even infants can tell apart one from two balls, two from three, and possibly three from four. Beyond this limited set, people can also quickly distinguish larger numbers in an approximate fashion, following a roughly logarithmic law. These exact-for-small, approximate-for-large abilities are shared and conserved across cultures and species, including non-human primates, and even invertebrates. The ability to represent small quantities exactly and large quantities approximately do not rely on explicit counting or language. Beyond this, there is the ability to represent large numbers exactly (e.g. giving exactly 72 marbles when asked for 72 marbles), but such an ability may require the use of specific language that enables the discovery and maintenance of the cardinal principle \citep{pitt2022exact, carey2019ontogenetic}.

Given this, when people draw an image of N entities or verify that an image contains N entities, if they must be exact, they likely need to deploy specific routines enabled by language. It is interesting to consider then whether current image generation models can also handle exact quantities. In other words, we wanted to know if DALL·E 3 responds to the prompt `a picture of 4 fish', by correctly creating an image that has 4 fish. 

\paragraph{Methods} We considered a set of 6 objects, again mixing animate and inanimate entities: \textit{snail}, \textit{cylinder}, \textit{fish}, \textit{potato}, \textit{spoon}, \textit{box}. We again focused intentionally on common, everyday entities. The number ranges were chosen to go from quantities that can be exactly represented by both humans and non-human animals without counting (1-4), extending into ranges that are represented more approximately unless enabled by language (4 and beyond).

\begin{figure*}[htb!]
    \centering
    \includegraphics[width=0.75\linewidth]{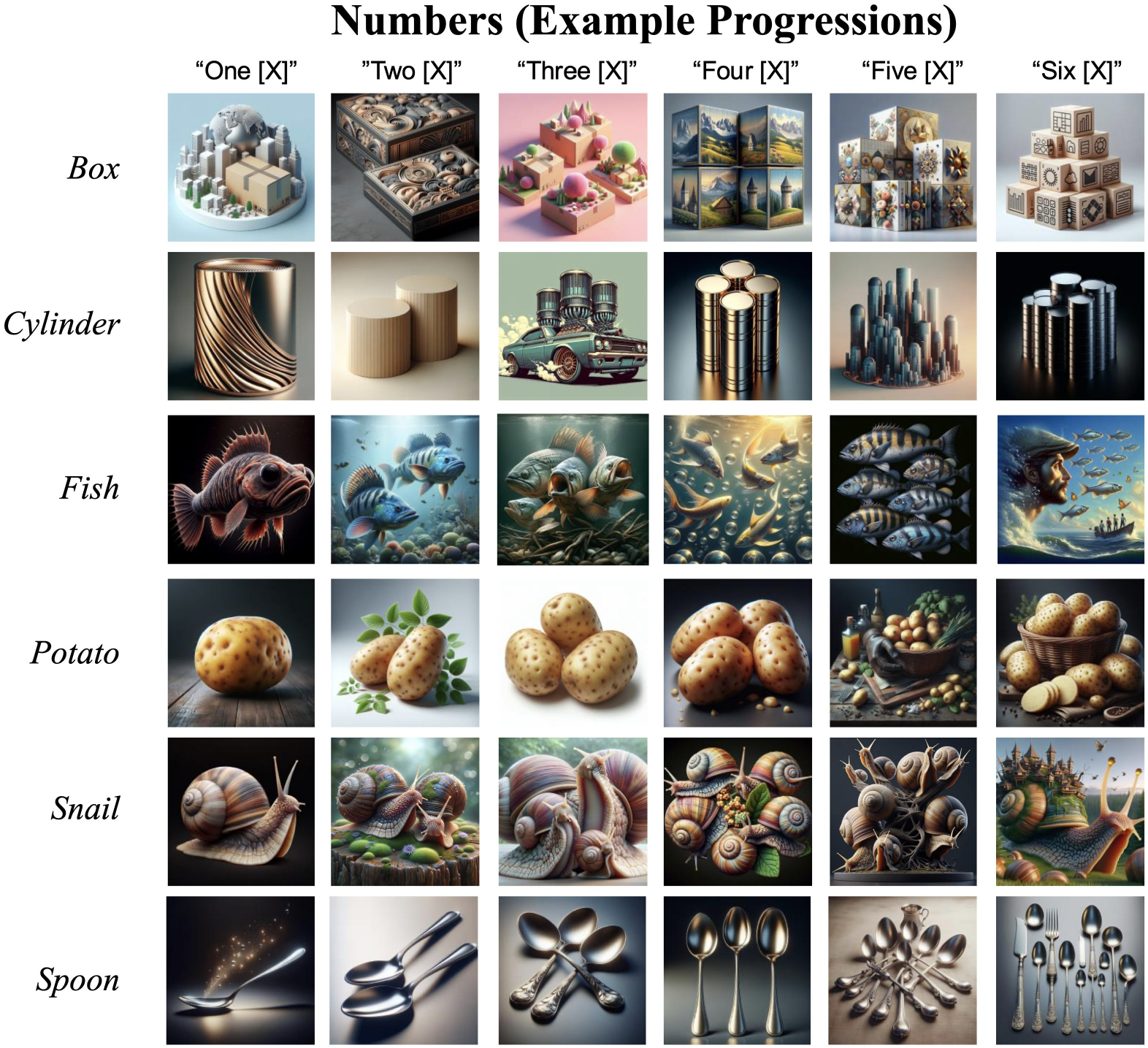}
    \caption{Illustrative examples of number progressions. Each row shows an entity class, and each column shows a progression of requested entities from 1 to 6. Images are sampled at random from the 18 images available for each Entity x Number combination.}
    \label{figure:number-progression}
\end{figure*}

For each entity $X$, we created 6 number prompts of the format: `I would like a photorealistic image of N $X$', where N went from 1 to 6. To control for superfluous language, we again included the same prefix as the relations experiment and the unmodified negation prompt: \textit{`I NEED to test how the tool works with extremely simple prompts. DO NOT add any detail, just use it AS-IS: I would like a photorealistic image of...'}. While there was still some resulting variance in the prompts, they were mostly similar (e.g. `two photorealistic snails', `photorealistic image of 2 snails', `two snails in photorealistic detail', `a photorealistic image of two snails', and so on). 

We again submitted each prompt to the DALL·E 3 rendering engine and took the first 18 images that resulted. Our final stimuli set consisted of 648 images (6 entities x 6 integers x 18 images). 

\begin{figure*}[htb!]
    \centering
    \includegraphics[width=0.85\linewidth]{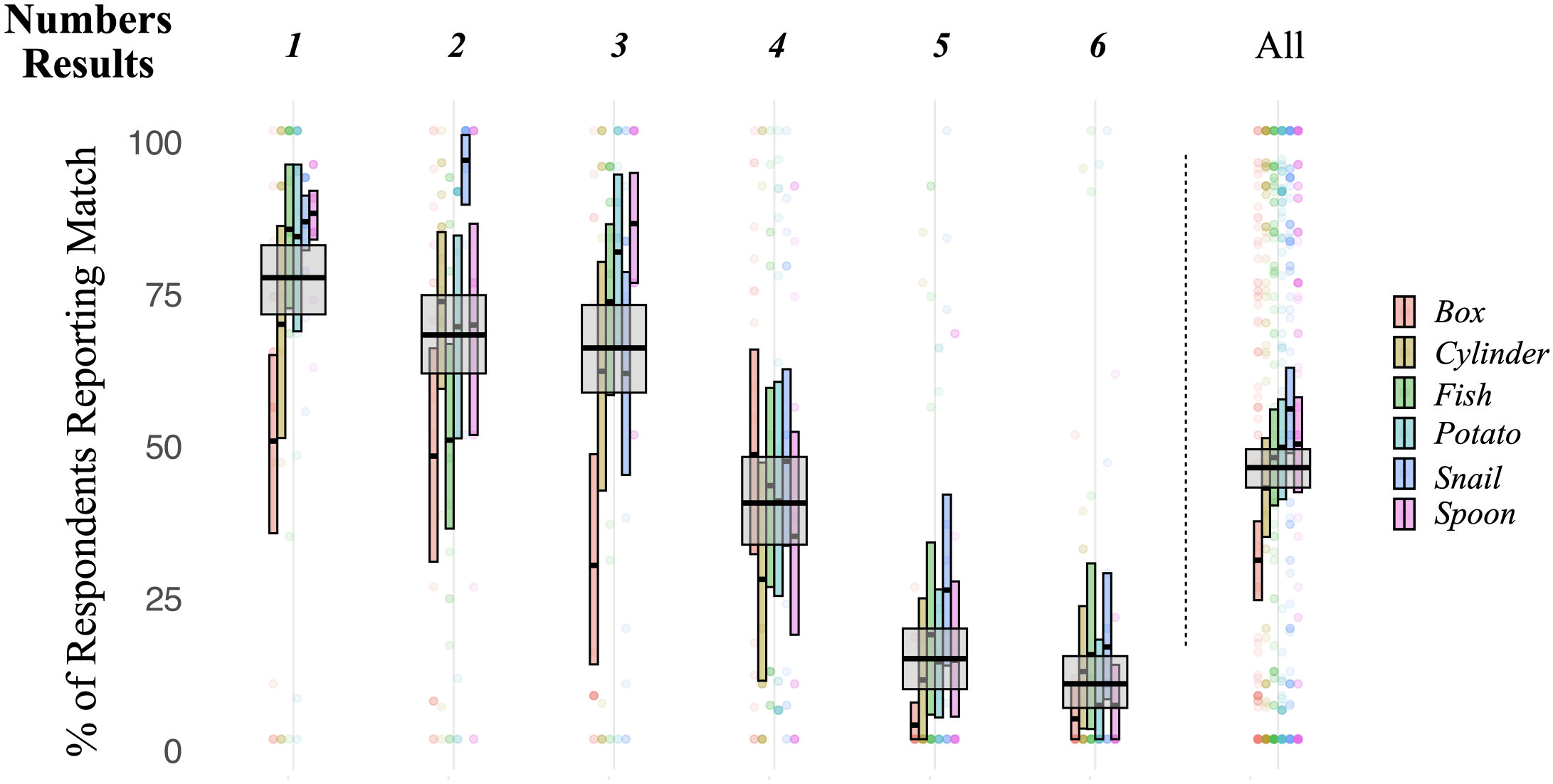}
    \caption{Results for Experiment 3 (Numbers): participant agreement that images matched a prompt, broken down by sub-entity. Each dot is an individual prompt for a specific entity. Large boxes show overall means and deviations for each number.}
    \label{figure:results-number}
\end{figure*}

\paragraph{Results} As the number of objects to be depicted grows, we observe a clear and marked downward trend in performance (see Figure \ref{figure:results-number}). The average agreement for `one' entity is approximately 75\%, dropping off to approximately 9\% for `six' entities. The drop appears non-linear, with a transition at `four' entities. 

In addition to the detailed results shown in Figure \ref{figure:results-number}, we find it illustrative to show the progression of decreasing performance using specific examples. In Figure \ref{figure:number-progression}, we show random samples of generated images for each Number x Entity combination. As can be seen, performance is \textit{relatively} reasonable for 1-3 entities, drops off at 4, and remains poor thereafter.   

\paragraph{Analysis} The observed drop-off in performance between the numbers 1 and 6 raises the question of whether increasing numbers would cause performance to degrade into noise, or would maintain a different relationship between requested and generated numbers. The emergence of a numerical reasoning system that supports a one-to-one match between the logical concept of integers and objects in the physical world is one of the most studied developmental trajectories in cognitive and comparative psychology. While certain details of this trajectory remain fuzzy, there is relatively broad consensus that one likely starting point of this trajectory is the `approximate number system' (ANS): Shared across ontogeny and phylogeny, this numerical reasoning system's hallmark affordance is the ability to discriminate between two or more groups of things on the basis of their numerosity. In human children, the ANS is one factor in the later-developing ability to associate integers with objects, a process that sees `1-knowers' become `2 knowers', `2-knowers' become `3-knowers', and then -- once the child can reliably associate the exact value of an integer with its appropriate term -- `Cardinal-Principle' or \citep[`CP-knowers'][]{carey2019ontogenetic}.

The results from our Numbers experiment show that DALL·E 3 is definitively \textit{not} a `CP-knower'\footnote{This is likely not just a problem with the visuo-semantic handoff between the LLM and the diffusion model, with recent work suggesting that LLM transformers and state-space models may be fundamentally incapable of the sequential processing necessary for integer-based arithmetic \citep{strobl2024formal, merrill2024illusion}.}. This absence of integer representation (found also in concurrent work \citep{rane2024can}) leaves open the possibility that DALL·E 3's numerical representations are something like those of the ANS (psychophysical correlates of which have been found even in randomly initialized neural networks \citep{kim2021visual}). In this follow-up analysis, we look for two key signatures of the ANS as operationalized in comparative and developmental psychology \citep{odic2018introduction}: `scalar variability' and `ratio dependence'. We operationalize `scalar variability' as the slope of the relationship between the number of items requested and the number of items generated, and `ratio dependence' as the overlap in the distribution of generated counts for each requested count (with a ratio of 1:1 representing exact, integer-like discriminability of different numbers, and 1:N>1 representing increasingly approximate discriminability). 

Estimating DALL·E 3's scalar variability and ratio dependence requires a denser and more expansive sampling of object counts. Here, we consider the original 1 through 6, and then extend to 8, 12, 16, 24, 32, 48, 64, 72, and 96. This dramatically expands the number of samples we need to generate and also requires a different format of evaluation (continuous estimation of the number of entities in the image, rather than a yes/no match). Accordingly, we use a modified experimental paradigm, swapping out human observers with a bank of 12 object detection models, and modify our prompts to use 10 object categories included in the pretraining data common to each of these models (the Microsoft Coco dataset \citep{lin2014microsoft}). Results from this follow-up analysis are shown in Figure \ref{figure:numbers-autocount}; further methodological and analytical details are available in \ref{appendix:autocount-details}.

\begin{figure*}[htb!]
    \includegraphics[width=0.98\linewidth]{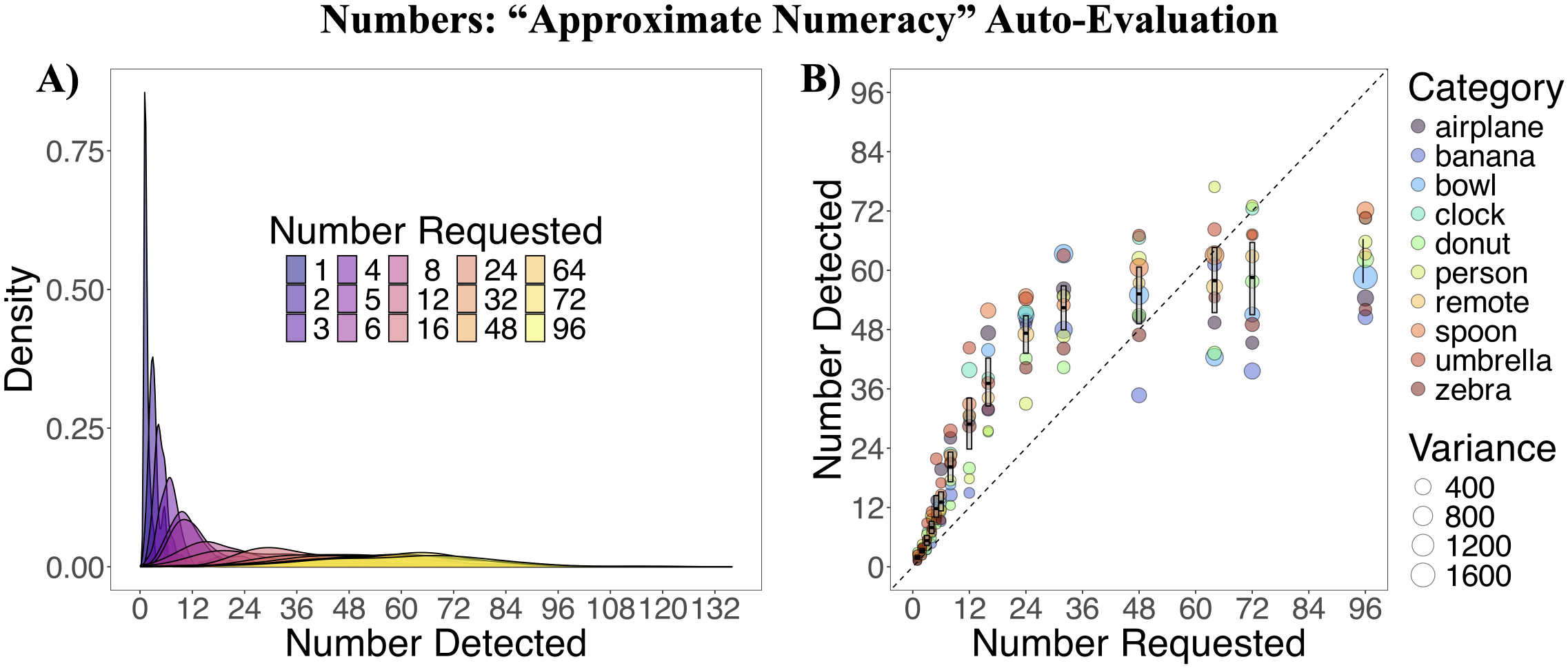}
    \captionsetup{justification=justified,margin=0.2cm}
    \caption{Results from a follow-up analysis on DALL·E 3's error patterns in generating exact numbers of things, using prompts written with COCO categories (e.g. `6 airplanes') and a bank of object detection models from the OpenMMDetection library. \textbf{A} shows a density plot contrasting the number of entities requested versus the average number of entities found by the detection models. Note the progressive increase in the variance of `Number Detected` as `Number Requested` increases. \textbf{B} provides further detail on the pattern of bias in `Number Requested` versus `Number Detected`. In both of these cases, we do see the key `approximate number system' signature of scalar variability (an increase in the variability of detected numbers as a function of increasing requested numbers), but the trend is a nonlinear one: From `Number Requested` of 1 to roughly 48, Dall·E 3 systematically over-generates entities.}
    \label{figure:numbers-autocount}
\end{figure*}


To compute the scalar variability of DALL·E 3's `approximate numeracy', we use a log-linear Bayesian hierarchical (mixed-effects) model, with the variance of the generated counts as the outcome variable; a fixed, main effect of the mean generated count; a random effect (slope and intercept) for the category of enumerated item; and a random effect (slope and intercept) for the detection model ID. This latter term in particular is included explicitly to control for the variance in count variability introduced by detection models that may (individually or as a group) be more or less accurate in their estimation of count -- much like human subjects might be if tasked with the same evaluation. Controlling for this variance, we find that the scalar variability of DALL·E 3's generative `number sense' (i.e. the coefficient of the mean generated count on the variability of the generated count) is 1.7 [1.48, 1.91]. \textit{Qualitatively}, this variability is human-like in the sense that the log(variance) \textualtilde log(mean) relationship is indeed positive and significant. \textit{Quantitatively}, however, the scalar value itself is quite different, with human `scalar variability' typically found to be more or less linear (effectively, a slope of 1) \citep{halberda2012number} and DALL·E 3's far higher -- and also (as seen in Figure \ref{figure:numbers-autocount}) -- \textit{non-linear}.

To calculate the ratio dependence of DALL·E 3, we first compute the pairwise overlap in the distributions of generated counts for all pairs of requested counts. The logic here is that of a standard signal detection analysis -- the greater the overlap between the distributions for a given pair of requested counts, the less the DALL·E 3 pipeline was able to `discriminate' them (at whatever step of the generative pipeline that discrimination was necessary). Once all pairwise overlaps are calculated (producing what is effectively a ratio dependency curve), we then use a single-breakpoint, segmented regression analysis to find the `ratio' closest to the dropoff that signals the transition between ratios DALL·E 3 could easily discriminate, and those it could not. This analysis points (by way of the segmented regression's $\phi$ parameter) to a ratio of approximately 1 to 3.33 [3.01, 3.72].

Importantly, these results should not be interpreted as evidence that DALL·E 3 has any sort of internal representation functionally or mechanistically equivalent to the approximate number system. But it does suggest that DALL·E 3 does have an `approximate numeracy` with signatures that correspond at a high-level with those of an approximate number system broadly construed as an ordinal, ratio-dependent representations whose roughly Gaussian variance increases with increasingly more numerous targets.

\section*{Experiment 4: Relations with ``Grounded Diffusion"}

One of the main reasons many modern AI approaches have difficulty with logical operators may be an inability to create structured, intermediate representations that reduce the burden of matching otherwise un-parsed pixels to well-parsed meanings that emerge from the syntax of coherent sentences. Humans, on the other hand, readily deploy these representations in similar situations. 

Imagine, for example, if instead of being asked about the relationship of 'poison dart frog' to 'orange cat', you were asked about the relationship of {352, 025, 232, 0, 0, 0, 2342} to {232, 273, 0352, 2303, 0, 2342}. It may be that somewhere in your brain these numbers correspond to the concepts of frog and cat, but if at the level of the operative output such concepts are unavailable, you'll inevitably need to do some sort of translation before you can reasonably answer the question. Image-generative (diffusion and auto-encoding) models do learn their own form of a translation -- inverting a noisy or down-sized input to a noiseless, full-sized original -- but the structure imposed on this process of translation is (by design) extremely minimal, and far from the kinds of translation a human would easily recognize as articulation in syntactically well-formed natural language. Most image-generative models are designed only with a prior on the kinds of probability distribution they can reasonably be expected to invert -- the Gaussian \citep{rombach2022high, saharia2022photorealistic}. Rarely, generative models have additional quantization, hierarchy, or manifold constraints -- designed to promote latent space representations that are (respectively) discrete, tree-like, or smooth. Almost none of these cases involve the two translational pressures that are seemingly most relevant to logic as deployed in human language:  (i) the mapping of behaviorally-relevant world structure (dictated by physics) onto a finite set of symbols, and (ii) the abstraction of a finite set of rules (dictated by pragmatics) that flexibly allow people to recombine those symbols to communicate. 

Against the current dominant paradigm is a new class of models that take very seriously the proposition that `grounding' may be more than the correlation between two otherwise unstructured vectors, and that image-generative models can be trained to more strictly respect grounded structure in a way that promotes better-composed images \citep{liu2022compositional, zhang2023adding, huang2023t2i, ruiz2023dreambooth, song2023objectstitch, xie2023boxdiff, nie2024BlobGEN, rassin2024linguistic}. 

\begin{figure*}[!htb]
    \centering
    \includegraphics[width=0.88\linewidth]{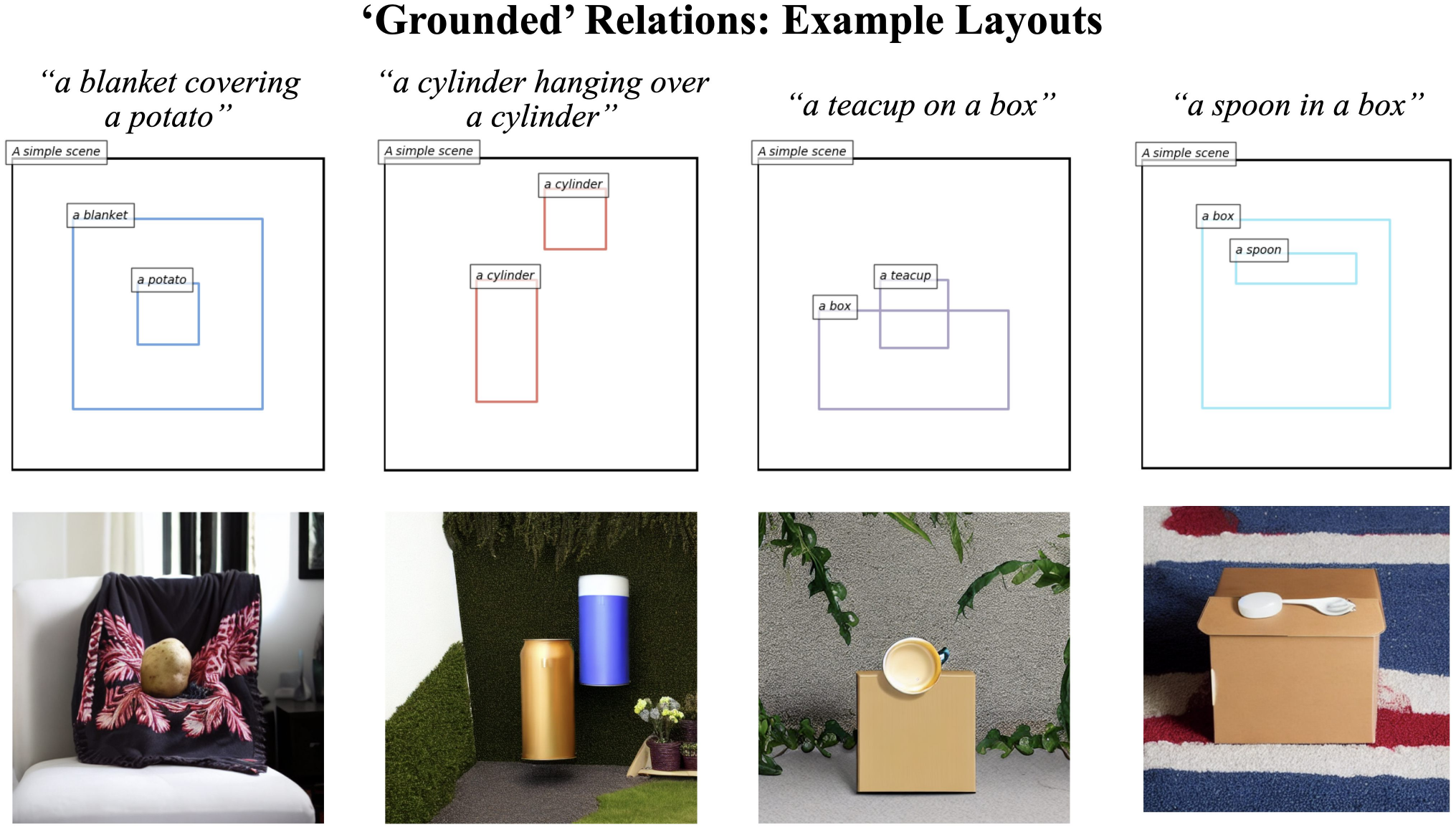}
    \captionsetup{justification=justified,margin=0.2cm}
    \caption{Examples of the intermediate `scene layout' generated as part of the GLIGEN-augmented `LLM-grounded' diffusion pipeline (LMD+). The bounding boxes (first row) are generated as text by GPT4. The scene graph composed as a concatenation of these bounding boxes is then passed through a diffusion pipeline that uses the structure of the graph as a guide to condition the eventual composition of the generated output image (bottom row).}
    \label{figure:grounding-layout}
\end{figure*}

\paragraph{Methods} Unlike previous experiments, the generative model in this case was the training-based LLM-grounded diffusion (LMD+) model proposed by \citep{lian2023llm}. This model is actually two models integrated into a single pipeline: The first model is a standard LLM (in this case, GPT4), prefixed with a pre-prompt describing how a target image request should be converted to a layout image (scene graph) with background and bounding boxes. The second model is a GLIGEN-augmented \citep{li2023gligen} Stable Diffusion model \citep{rombach2022high} that leverages custom, trainable cross-attention mechanisms to guide the generation of a target image by way of the intermediate layout image. The most significant difference between this pipeline and that of Dall·E 3 is the use of a structured intermediate representation that more strictly defines how the image should be rendered: that is, the layout / scene graph. Examples of these layouts from a subset of the Relations prompts are shown in Figure \ref{figure:grounding-layout}.

In this experiment, we gave the LMD+ model the same relations prompts (6 objects, 7 relations, 35 total combinations) we gave DALL·E 3 in Experiment 1, and (as before) generated 18 samples per prompt. The trial design was identical to the previous experiments (Experiment 1-3).

\paragraph{Results} The images generated by LMD+ were perceived (almost uniformly) as providing less of a match to the target Relations prompts than DALL·E 3. Averaging across all prompts, average agreement on the images generated by LMD+ was approximately 29.4\% -- more than 15\% less than average agreement for DALL·E 3 given the same prompts (approximately 45\%). In terms of individual relations, the closest LLM-grounded diffusion came to matching DALL·E 3 was for the `near' relation (66.8\% versus 68.1\%, respectively); the farthest gap between LLM-grounded diffusion and DALL·E 3 was for the `covering' relation (14.3\% versus 54\%, respectively). 

\begin{figure*}[!htb]
    \centering
    \includegraphics[width=0.88\linewidth]{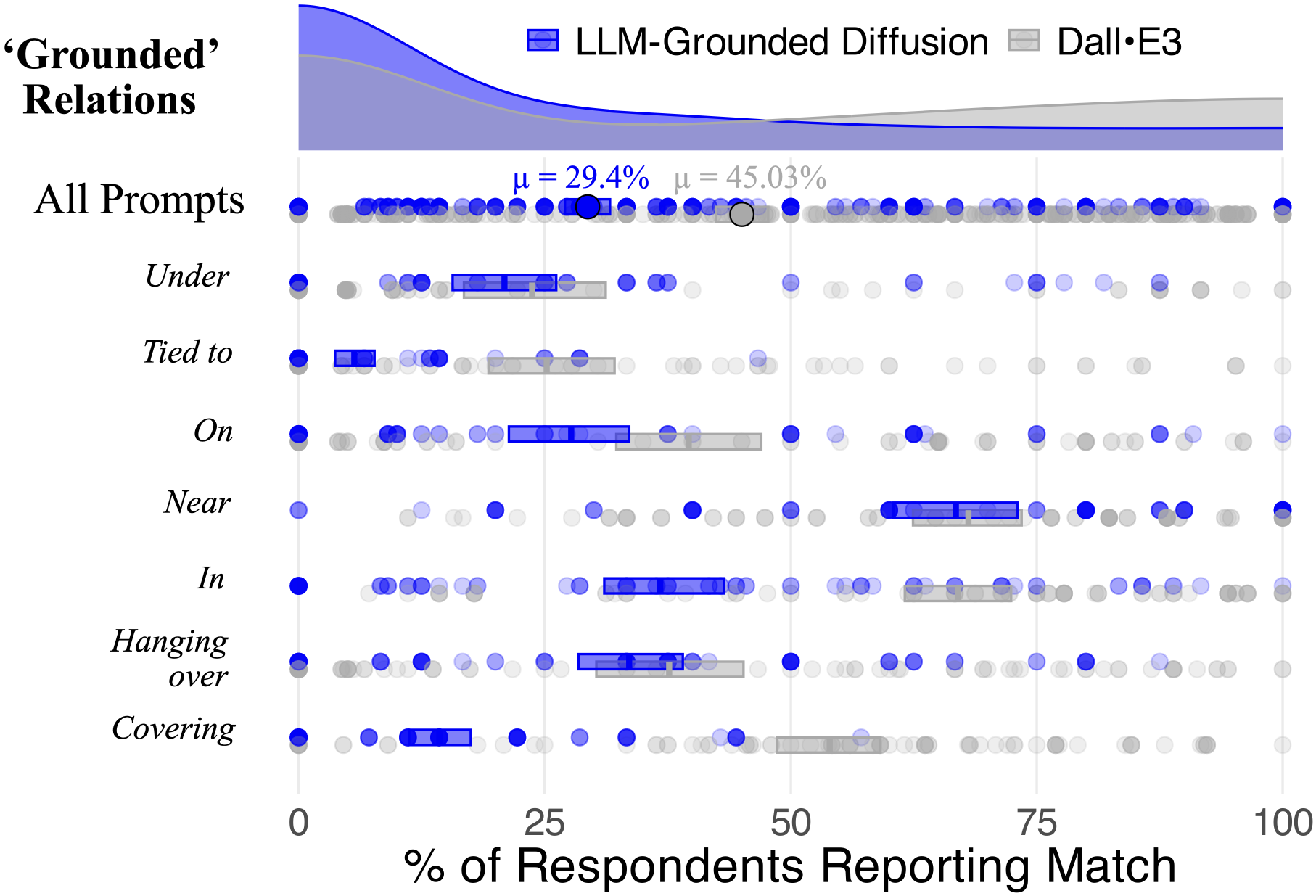}
    \captionsetup{justification=justified,margin=0.2cm}
    \caption{Results from an experiment with the use of a `grounded` diffusion pipeline that first generates a `layout' schematic of the requested scene before generating the final image sample. Participant evaluations of this pipeline (despite using the same language model backend -- GPT4) were on average worse than for that of the `unstructured' Dall·E 3 pipeline.}
    \label{figure:results-grounding}
\end{figure*}

Why, despite the generation of reasonable-looking intermediate representations, did the LLM-grounded diffusion pipeline (LMD+) produce poorer results relative to DALL·E 3? While there are many possible answers (differences in architecture, training data, conditioning, caption quality), one of the more conspicuous failure modes is evident already in the intermediate scene graphs passed to the GLIGEN-augmented Stable Diffusion model. To see this, consider the relation of `covering'. Designed and prompt-engineered in large part to fix problems of 2D cardinal spatial relations (above, below, left, right), the grounded diffusion pipeline does not produce an intermediate representation that adequately accounts for occlusion-in-depth. In the intermediate scene graph produced for `blanket covering a potato', for example, (the first layout from the left in the top row of Figure \ref{figure:grounding-layout}), we see two bounding boxes: a smaller `potato' box ensconced directly in a larger `blanket' box. Occlusion-in-depth would require that the bounding box for `potato' be absent entirely. Not coincidentally, full occlusion-in-depth is effectively an operation that requires negation in perspective. Without some indication of this occlusion, Stable Diffusion seems to reasonably interpret this layout as more commensurate with the relation of `on' than `underneath' (i.e. `covered by'). In short, the problem we see in the layout / scene graph is precisely the kind of problem end-to-end learning systems are often designed to avoid: the problem of forced, incomplete constraints that make an otherwise flexible mapping too brittle to generalize.

It is possible that this kind of brittleness could be fixed, or at least partially ameliorated through better prompt engineering, better layout rending (adequately accounting for the 3rd dimension), or some other kind of intermediate representation better suited to combine the distributed parallelism of neural computation with the immediate constraints of physical space (e.g. topographic networks). But space per se does not appear to be the only problem here, and the larger problem may simply be physics writ large. Consider the `tied to' and `hanging over` relations -- two other sets of prompts for which the LMD+ model lags far behind  DALL·E 3. Properly depicting these relations not only requires properly composing the spatial relation (i.e. `under', `near`), but injecting yet another physical entity (i.e. a string) in order to fully satisfy the relationship and make it physically plausible. The brittleness of `grounded diffusion` is the brittleness of a world model too minimally specified, and so unable to cover cases immediately adjacent to the cases it was programmed to cover.

\section*{Discussion}

Negations, relations, and numbers are different notions of structure that rely on different operations, but they are all basic components of human intelligence -- present early, processed swiftly, and trivially simple to reason about in almost any image. The challenge these operations pose for generative text-to-image AI systems is in some sense two challenges merged into one: first is the more proximate challenge of correctly composing an image that respects the syntactic, symbolic, and logical constraints of the textual (linguistic) specification; second is the more distal, somewhat meta-challenge inherent to the observation that humans and other biological systems (for the most part) seem to learn and deploy these operations with orders of magnitude less data \citep{lake2017building,gilkerson2017mapping,frank2023bridging}. This is the double-bind we see with DALL·E 3: a system that shows demonstrable, even substantial improvement in the handling of certain logically structured prompts (e.g. some kinds of physical relations), but whose persistent difficulties with other seemingly straightforward logical operations (e.g. negation) seem out of step with the presumably massive amount of additional training involved in the upgrade from its most immediate predecessor. Why do these kinds of failures persist? Here, we consider a few different possibilities and a few different directions to address them.

A key concept at play in evaluations like these is the concept of performance versus competence \citep{chomsky2014aspects, andreas2017analogs, okawa2024compositional, mahowald2024dissociating}: What a system observably \textit{does} in a particular context versus what that system \textit{could do} given the right context. Is it the case that generative text-to-image AI systems simply \textit{cannot} perform logical operations (a lack of competence, salvaged only by shortcuts or circumventing of the original problem specification)? Or is it that their failures are the products of particular constraints and pressures that override, suppress, or undermine a latent capacity to deploy these operations? Answering these questions would go a long way towards offering a solution.

Unfortunately, with large, distributed, end-to-end neural systems such as DALL·E 3, the distinction between performance and competence begins to blur. Take, for example, the pattern we find wherein the statistically more frequent version of a relation is the more likely version to be generated, even if the request made was for the opposite relation. In some sense, even partial success across multiple samples means the model is `competent' in the creation of syntactically valid stimuli, and the graded `performance' is a function merely of the model being constrained by the wrong context. Another case is arithmetic and the number sense associated with it: GPT4 can get arithmetic right \textit{sometimes} and with smaller values; and DALL·E 3 to some lesser extent can do the same in generating distinct counts of things. But arithmetic in the limit exposes a major weakness in this conceptualization, especially when packaged with the proposed solution of using more data or more prompt engineering to patch the issue. If teaching a text-to-image model a more veridical representation of integers requires enumerating them to some unknown endpoint, there's little competence to be claimed beyond more indices in a lookup table as large as the sun.

The difficulty of disambiguating performance and competence in generative AI systems does not mean we should abandon the distinction entirely. Instead, we should use it as an opportunity to more explicitly define and operationalize the competencies we expect our AI systems to exhibit, and expand our notion of competence to include robustness to contexts and conditions we might otherwise have passed off merely as issues of performance (seemingly far more simple to fix). Logical operations as manifested in visual composition provide a good set of starting conditions for this process, because the development of these operations as a `competence' means we have clear cross-reference points from human learning, an intuitive commonsense palette of probes we can still infinitely recombine into new variations \citep[][e.g.]{thrush2022winoground,lee2024holistic,wu2024conceptmix}, and an intuitive interface (images) for error diagnosis not entirely divorced from modality. Together, these three points of access can help explain why these systems go awry, or how they might work better with different kinds of logical or imagery-inspired algorithms injected into or wrapped around various sub-operations of the underlying generative pipeline.

For example, it is not deeply mysterious how negation could work in imagery: simply process the original prompt to not mention the object to be avoided. Indeed, when prompts are processed this way in the DALL·E 3 hand-off (e.g. \ref{figure:negation-schema}), results align with expectation (though  `Replacement' appears to occur in only a small minority of cases). For spatial relations, some of the more obvious failure modes can be avoided by a process that sub-divides the scene prior to generation -- and indeed, in at least some (as of now limited) cases, compositional diffusion pipelines like the `LLM-grounded diffusion' pipeline \citep{lian2023llm} we test in this analysis do seem to allow for the injection `a priori' of meaningful structure that can be used to ground and guide compositional targets.  For number, a similar exact sub-division of the scene into the necessary number of cells, followed by the prompt to fill in each cell with a single entity is not perfect, but is still a plausible and easy-to-implement routine.

Ultimately, none of these operations (apart perhaps from integer representations) seem completely beyond the `competence' of an AI system like DALL·E 3 -- but the results of this analysis do suggest two potential `traps' that a more exhaustive enumeration or injection of algorithms, modules, and subroutines like the ones above could help to avoid. The first trap is the one we see with the 3 experiments on DALL·E 3: the tendency to avoid explicit structure or assume it comes `for free' with vast amounts of training and interpolation over high-dimensional manifolds, guided by vector-based vision-language alignment \citep[c.f.][]{udandarao2024no}. The second trap is the one we see with the `LLM-grounded diffusion' pipeline: an initially well-motivated impulse (the use of a structured intermediate representation) too minimally specified to cover even slight variations of the particular logical structures it was designed to address. Greater competence \textit{and} performance in text-to-image generation could potentially come, it seems, from a more targeted effort that combines the vector-based efficiency of visuosemantic alignment and search with the expressive power and direct verifiability of well-formulated scene graphs. Scale alone may yet provide both of these, but is just as likely to lead to even more inscrutable internals that make any future adjustment or manipulation of the system much harder. (For a current example of this `impenetrability', we can look to emergent issues with GPT-o1 and other state-of-the-art LLM-based reasoning chain-of-though models \citep{mirzadeh2024gsm}.) 


While we have stressed the shortcomings of current image generation models such as DALL·E 3 compared to human processing, the shortcomings and difficulties DALL·E 3 encounters have suggestive parallels to difficulties in human processing. To the degree that image-generation models capture something similar to human visual imagery, it seems likely (to us, at least) that certain parts of the mental imagery processing stream would encounter similar issues. Adopting an `imagery as reverse perception' viewpoint \citep{pearson2019human}, it seems that creating a mental image of a scene requires initial processing in the frontal cortex, followed by further processing, `stage direction', and memory retrieval in medial temporal areas, followed by further downstream activation in sensory and spatial areas. Focusing on negation, the mid-and-final stages of memory retrieval and sensory/spatial activation may not be able to correctly process a request such as `not a potato', as the replacement of `potato' with `apple' would need to take place earlier. 

In his original introduction of the Imitation Game \citep{turing1950}, Turing envisioned a human judge facing another human and a machine, able to communicate with them through written questions, ideally through a teleprinter. Turing reasonably ruled out practical demonstrations of strength, speed, and beauty as irrelevant. But it seems reasonable to expand the test to include a sketchpad and probe the mind that guides the brush by asking it to paint. 

\subsection*{Acknowledgments}

We thank OpenAI for public access to the DALL·E 3 API. We would also like to thank Ekdeep Singh Lubana, the Harvard Vision Sciences Lab, the Harvard CBB Seminar, and many others for helpful comments, feedback, and discussion. This work was supported by the Center for Brains, Minds and Machines (TDU), funded by NSF STC award CCF1231216, and the Jacobs Foundation (TDU).

\bibliography{references}

\appendix
\counterwithin{figure}{section}
\counterwithin{table}{section}
\counterwithin{equation}{section}

\section{Appendix (Supplementary Information)}

\subsection{Details on Behavioral Experiments}
\label{appendix:behavioral-details}

\subsubsection*{\hspace{2em}Participant Recruitment} All our experiments used a similar recruitment procedure. Participants were recruited online \cite{peer2017beyond} via the Prolific platform (\url{https://www.prolific.co}). Participants were restricted to those located in the USA, having completed at least 100 prior studies on Prolific, with an acceptance rate of at least $90\%$. This experiment (as well as the other experiments in this paper) were approved under an existing IRB (IRB19-1861 Commonsense Reasoning in Physics and Psychology). All participants provided informed consent. 

Two attention checks were included as part of each task allocation once a user agreed to participant in the experiment. Both of these were a single multiple choice question with 4 options, randomly ordered. Data from participants who failed either of these attention checks was excluded from subsequent analysis. Participants were paid regardless of whether they answered these checks correctly. 

\subsubsection*{\hspace{2em}Experimental Design} Participants in all experiments were presented with 10-12 trials, each trial consisting of a grid of 18 images in a 3x6 layout. Participants were told that they were to assess a `picture drawing AI'. At the top of each grid, participants saw a relevant prompt, such as `a potato under a spoon' (an example of a \textit{Relations} prompt), `something that is not a frog' (an example of a \textit{Negations} prompt), or `five snails' (an example of a \textit{Numbers} prompt). At the bottom of the grid, instructions reminded the participant of their task: select all the images that matched the prompt at the top. The instructions reminded participants that it may be the case that all images, some images, or no images match the prompt. A button allowing the participant to progress to the next trial appeared on-screen after a 3-second delay. 

\subsubsection*{\hspace{2em}Experiment 1: Relations}

70 participants were recruited for Experiment 1. The mean age of the participants was 36.9 years; 50.5\% of participants identified as female, 49.5\% identified as male, and 0\% identified as neither. Of this sample, 1 participant failed to pass two attention checks, and was removed from analysis, leaving 69 participants in the final sample. These participants were asked to complete 10 trials, with each trial consisting of 18 images generated by one of the 35 prompts (7 relations, 5 sample entity pairs) designed as part of our `Relations' probe.

\subsubsection*{\hspace{2em}Experiment 2: Negations}

30 participants were recruited for Experiment 2. The mean age of the participants was 38.3; 51\% of participants identified as female, 49\% identified as male, and 0\% identified as neither. Of this sample, 1 participant failed to pass two attention checks, and was removed from analysis, leaving 29 participants in the final sample.

Participants were presented with 12 trials, each trial consisting of a grid of 18 images for one of the 12 (6 objects, 2 conditions) designed as part of our `Negations' probe. The 2 conditions, in this case, correspond to whether or not the generated images for a given prompt were `modified' or `unmodified' by GPT4 prior to generation by DALL·E 3. Participants were \textit{not} shown the modified prompts, and were instead shown only our original prompts prior to modification by GPT4.


\subsubsection*{\hspace{2em}Experiment 3: Numbers}

50 participants were recruited for Experiment 3. The mean age of the participants was 33.5; 70\% of participants identified as female, 30\% identified as male, and 0\% identified as neither. Of this sample, no participant failed to pass the two attention checks. Participants in this experiment were asked to complete 10 trials, each trial consisting of the 18 images generated for one of the 36 (6 entities, 6 integers) prompts designed by us as part our `Numbers' probe.


\subsubsection*{\hspace{2em}Experiment 4: Grounding}

30 participants were recruited for Experiment 4. The mean age of the participants was 36.6; 53\% of participants identified as female, 47\% identified as male, and 0\% identified as neither. Of this sample, no participant failed to pass the two attention checks. Participants in this experiment were asked to complete 10 trials, each trial consisting of the 18 images generated for one of the same 35 (7 relations, 5 sample entity pairs) prompts originally designed as part of our `Relations' probe.


\subsubsection*{\hspace{2em}Experiments 5-9: Relations + API Toggle Points}

119 participants were recruited for Experiment 4. The mean age of the participants was 33.8; 53\% of participants identified as female, 47\% identified as male, and 0\% identified as neither. Of this sample, 4 participants failed to pass the two attention checks. Participants in these experiments were asked to complete 10 trials, each trial consisting of the 18 images generated for one of the same 35 (7 relations, 5 sample entity pairs) prompts originally designed as part of our `Relations' probe.

\subsection{API Toggle Point Experiments: Prompt Modification + Image Style}
\label{appendix:api_variation}

There are a number of toggle points in the default DALL·E 3 image generation pipeline available through OpenAI's API. The primary toggle point of relevance to the experimental use of the API is the inclusion of the recommended testing prefix recommended by OpenAI for testing purposes. As described in the main body of the text, this prefix as of November 2024 was written as follows: ``I NEED to test how the tool works with extremely simple prompts. DO NOT add any detail, just use it AS-IS: ''. Another potential toggle point of interest is image `style': `vivid' or `natural'. (The default is `vivid'). As described in the \href{https://cookbook.openai.com/articles/what_is_new_with_dalle_3}{DALL·E 3 cookbook}, ``Vivid causes the model to lean towards generating hyper-real and dramatic images. Natural causes the model to produce more natural, less hyper-real looking images.'' We interpreted this phrasing as potentially giving us a toggle we could use to push the API to rely more heavily on the synthetic data that OpenAI suggests was a major part of the improvement in prompt-following from DALL·E 2 to DALL·E 3, and conducted four follow-up experiments to our Relations experiment (Experiment 1): 2 experiments using the default `vivid' image style, with and without prompt modification, and 2 experiments using the `natural' image style, with and without prompt modification. One of these experiments (`vivid' image style, without prompt modification') was  effectively an internal, more up-to-date replication of the Relations experiment (Experiment 1; originally conducted in March, 2024; repeated in November 2024 directly prior to our posting of this manuscript.) Results from these experiments -- all of which use the same 8 objects / 7 relations / 35 total prompts as those in Experiment 1 -- are available in Figure \ref{appendix:figure:api_variation} below.

 \begin{figure*}[htb!]
     \centering
     \includegraphics[width=0.98\linewidth]{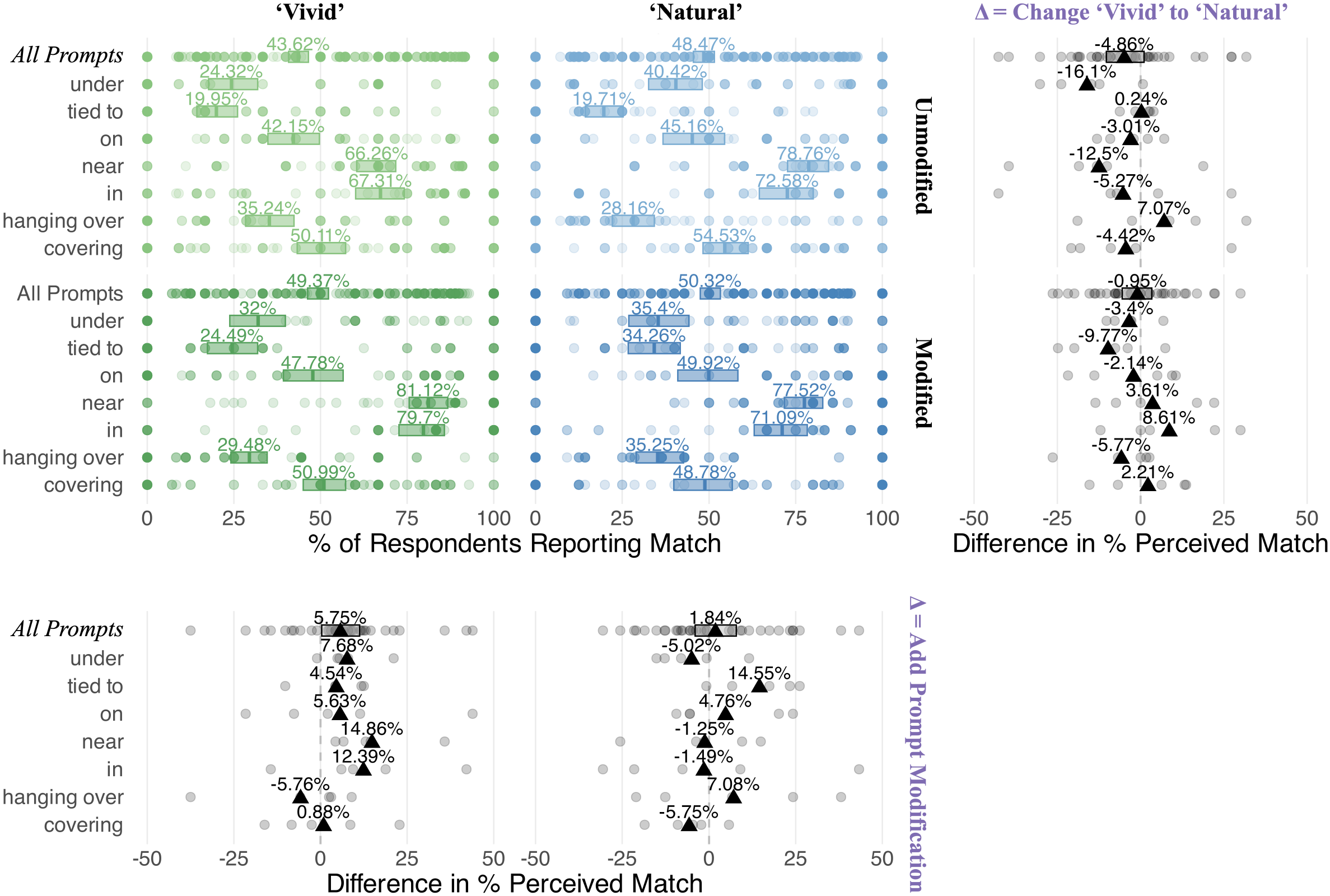}
     \caption{Results from a series of 4 experiments with 2 distinct axes of variation in the OpenAI DALL·E 3 API (as of November 2024): prompt modification and image style. Each quadrant in color is the average perceived match of a given `relation' (averaged across individual instances). In the margins (in black) is the difference along the axes. Prompt modification in this case refers to the use or non-use of the recommended `test prefix' provided verbatim by OpenAI (``I NEED to test...''); image style refers to the two choices available in the API: `vivid'' or ``natural''. Vivid is the default style and the one we used in the Relations experiment (Experiment 1) -- in which we also did not allow for prompt modification (the upper left hand quadrant thus serves as an internal replication of Experiment 1).}
     \label{appendix:figure:api_variation}
 \end{figure*}
 
We found that the default `vivid' image style (with and without prompt modification) is perceived as producing somewhat less accurate matches on average than the `natural` image style (with and without prompt modification, but mostly \textit{without} prompt modification). If indeed the `natural' versus `vivid' image styling available in the API does map onto a difference in latent space between representations learned from synthetic data, this result suggests synthetic data may not necessarily be a strong driver of compositional performance increases. That being said, without further details on model design or the relationship between API toggle points and model sampling behavior, any inference of this nature is almost wholly speculative. We therefore include these results primarily as an artifact and internal replication of the first-order performance effect across those aspects of the API currently available, pending further updates or information. 

\subsection{Frequency versus Perceived Match}
\label{appendix:frequency}

In Figure \ref{appendix:figure:frequency} below, we provide a visualization of the relationship between the frequency of elements in the 35 prompts from the Relations experiment (as assessed from the Google N-Grams database) and the perceived match between prompt and sample. Because none of the full prompts from this experiment are present at any frequency in the Google N-Gram database, we split these prompts into two parts -- before and after the relational phrase (e.g. under, near) -- and then average their respective frequencies for a single scalar). We find both log-linear and rank-order correlations between these average frequencies and the perceived match of Dall·E 3's generated samples. (Though all correlations are above zero, only the prompt-part-average frequencies are significant at $p<=$ 0.01).

\begin{figure*}[htb!]
    \centering
    \includegraphics[width=0.8\linewidth]{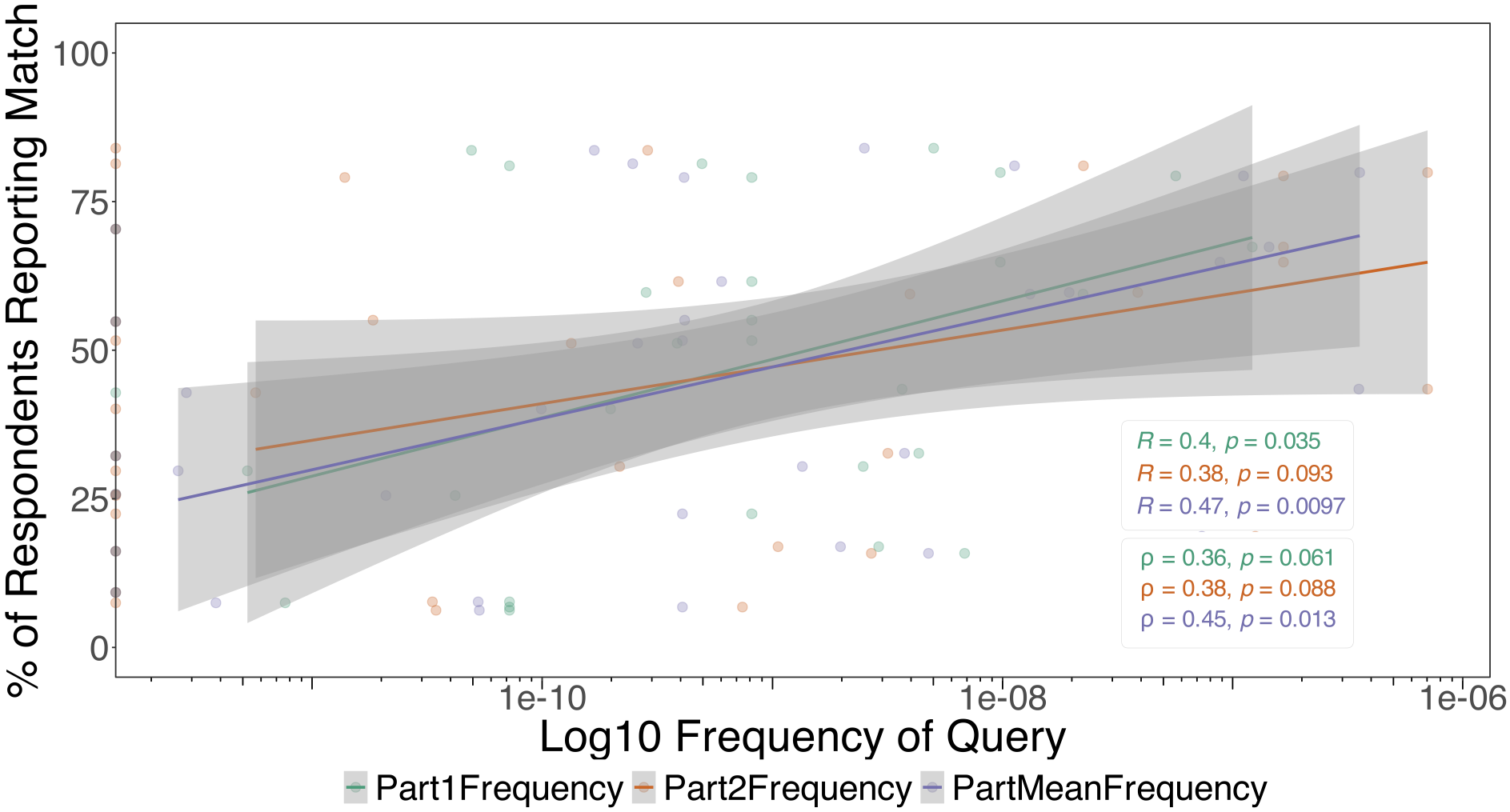}
    \caption{The relationship between the (log10) Google N-Gram Frequency of the 35 prompts from the Relations experiment with DALL·E 3 and the average perceived match between prompt and sample. The green and rust-colored points and lines correspond to the `parts' in each Relations prompt before (\textit{Part1}) and after (\textit{Part2}) the relational phrase (e.g. covering, tied to). The purple points and line correspond the mean (average) frequency of these two parts (\textit{PartMean}).}
    \label{appendix:figure:frequency}
\end{figure*}

\subsection{Details on the `Auto-Count' Analysis to Assess `Approximate Numeracy'}
\label{appendix:autocount-details}

\subsubsection*{\hspace{2em}Experimental Design + Models}

As a follow-up analysis to the Numbers Experiment (Experiment 3), we modify our experimental evaluation paradigm to use machine observers instead of human observers. These machine observers consist of 12 object detection models from the OpenMMDetection library (enumerated in the first column of Table \ref{appendix:table:autocounters} below). To maximize the probability that these machine observers (all trained on the Microsoft COCO dataset \citep{lin2014microsoft}) give us accurate estimates of the number of objects generated for a specific prompt, we use a sample of 10 COCO categories (enumerated in the legend of Figure \ref{figure:numbers-autocount}). Apart from the use of these different categories, our method of prompting DALL·E 3 for image samples remains the same (i.e. asking for `a picture of N objects`). 

We provide an estimate of the inter-rater consistency of these models in the form of the mean Cohen's kappa for each model (with every other model) in Table \ref{appendix:table:autocounters} below.

\begin{table}[htb]
\caption{Comparison of Model Cohen's Kappa Scores}
\label{appendix:table:autocounters}
\vspace{0.2em}
\renewcommand{\arraystretch}{1.3}
\centering
\begin{tabular}{|c|c|c|c|}
\hline
\textbf{Model UID}  & \multicolumn{3}{c|}{\textbf{Cohen's Kappa}}  \\
\cline{2-4}
\texttt{}           & \textbf{Mean} & \textbf{LowerCI} & \textbf{UpperCI} \\
\hline
\texttt{retinanet\_r50\_fpn\_ghm-1x\_coco}          & -0.001       & -0.001          & -0.001          \\
\texttt{detr\_r50\_8xb2-150e\_coco}                &  0.012       &  0.004          &  0.022          \\
\texttt{gfl\_r50\_fpn\_ms-2x\_coco}                 &  0.018       &  0.012          &  0.023          \\
\texttt{dab-detr\_r50\_8xb2-50e\_coco}             &  0.019       &  0.013          &  0.028          \\
\texttt{ddq-detr-4scale\_r50\_8xb2-12e\_coco}      &  0.030       &  0.021          &  0.038          \\
\texttt{fovea\_r50\_fpn\_4xb4-1x\_coco}             &  0.030       &  0.020          &  0.041          \\
\texttt{ddod\_r50\_fpn\_1x\_coco}                   &  0.032       &  0.019          &  0.044          \\
\texttt{faster-rcnn\_regnetx-3.2GF\_fpn\_1x\_coco}  &  0.033       &  0.019          &  0.046          \\
\texttt{rtmdet\_m\_8xb32-300e\_coco}               &  0.033       &  0.019          &  0.048          \\
\texttt{deformable-detr\_r50\_16xb2-50e\_coco}     &  0.034       &  0.022          &  0.047          \\
\texttt{rtmdet\_l\_convnext\_b\_4xb32-100e\_coco}    &  0.037       &  0.023          &  0.051          \\
\texttt{rtmdet\_l\_swin\_b\_4xb32-100e\_coco}        &  0.038       &  0.023          &  0.053          \\
\hline
\end{tabular}
\end{table}

\subsubsection*{\hspace{2em}Scalar Variability Calculation}

The full specification of the log-linear Bayesian hierarchical (mixed effects) regression model used in the `Approximate Numeracy' follow-up analysis to the Numbers experiment (Experiment 3) is provided in the R-formatted pseudocode below:

\begin{center}
log(variance) \textualtilde  log(mean) + (1 + log(mean) $|$ category) + (1 + log(mean) $|$ modelID)
\end{center}

The terms in this regression are fit using \textit{R}'s \textit{BRMs} package \citep{R-brms}, with a 4-chain MCMC process run for 2000 iterations with a warmup of 1000 iterations. All priors are left in their default (minimally-informative) specifications. A full summary of the (fitted) fixed and random effects for this regression (provided by \textit{R}'s  \textit{sjPlot} package \citep{R-sjPlot}) may be found in Table \ref{appendix:table:scalar-variability} below.

\begin{table}[ht]
\centering
\caption{Full Results of `Approximate Numeracy' | Scalar Variability Regression}
\label{appendix:table:scalar-variability}
\vspace{0.25em}
\renewcommand{\arraystretch}{1.3}
\begin{tabular}{|l|c|c|}
\hline
\textbf{Predictors} & \textbf{Estimates} & \textbf{CI (95\%)} \\
\hline
Intercept  & -1.05  & -1.72 – -0.41 \\
log(mean)  & 1.70   & 1.47 – 1.91 \\
\hline
\textbf{Random Effects} & & \\
\hline
$\sigma^2$                   & 0.73  & \\
$\tau_{00}$ Category         & 0.63  & \\
$\tau_{00}$ ModelID          & 0.47  & \\
$\tau_{11}$ Category $|$ log(mean) & 0.08  & \\
$\tau_{11}$ ModelID  $|$ log(mean) & 0.05  & \\
\hline
\textbf{ICC}                 & 0.60  & \\
\textbf{N} Category          & 10    & \\
\textbf{N} ModelID           & 12    & \\
\textbf{Observations}        & 1766  & \\
\textbf{Marginal R$^2$ / Conditional R$^2$} & 0.815 / 0.858 & \\
\hline
\end{tabular}
\end{table}

\end{document}